# Just aware enough:
# Evaluating awareness across artificial systems

**Nadine Meertens, Suet Lee, Ophelia Deroy**

Contact: nadine.meertens@lmu.de

## Abstract

Recent debates on artificial intelligence increasingly emphasise questions of AI consciousness and moral status, yet there remains little agreement on how such properties should be evaluated. In this paper, we argue that awareness offers a more productive and methodologically tractable alternative. We introduce a practical method for evaluating awareness across diverse artificial systems, where awareness is understood as encompassing a system's abilities to process, store and use information in the service of goal-directed action. Central to this approach is the claim that any evaluation aiming to capture the diversity of artificial systems must be domain-sensitive, deployable at any scale, multidimensional, and enable the prediction of task performance, while generalising to the level of abilities for the sake of comparison. Given these four desiderata, we outline a structured approach to evaluating and comparing awareness profiles across artificial systems with differing architectures, scales, and operational domains. By shifting the focus from artificial consciousness to being *just aware enough*, this approach aims to facilitate principled assessment, support design and oversight, and enable more constructive scientific and public discourse.

## 1. Introduction

From driving to work to grabbing items at the store or playing in an orchestra, we constantly rely on more than our conscious experience to guide our actions and responses. Contextual changes—in our environment, in others, and in ourselves—are registered and selected to orient our subsequent actions, often before or even without conscious attention or experience. To distinguish these capacities from consciousness, they are increasingly referred to as situational awareness, or simply awareness.

Awareness is being explored as a more neutral or workable alternative than AI consciousness for guiding AI and robotics research (Bacciu et al., 2025; Della Santina et al., 2025; Deroy et al., 2024; Evers et al., 2025; Li et al., 2025). On the one hand, it avoids premature or problematic attributions of consciousness to machines while remaining neutral on the possibility of AI consciousness in principle. On the other hand, it shifts the focus away from consciousness as a design goal.

In this paper, awareness is introduced not as a claim about the presence or usefulness of phenomenal consciousness, but as a way of characterising what artificial systems can select, integrate, and act upon within a given operational context. More precisely, it refers to a system's capabilities to process, store, retrieve and utilise information in the service of goal-directed action (Lee et al., 2026; Meertens, 2025). Awareness thus characterises context-sensitive information processing: the abilities systems have to selectively register and respond to changes in their environment, themselves, and other agents (Evers et al., 2025). As such, it functions not merely as a descriptive label, but as a notion intended to carry explanatory weight and to inform how artificial systems are designed and evaluated.

As artificial systems become ubiquitous—from credit scoring and medical diagnosis to wildfire response and goods delivery—they increasingly operate alongside humans and other agents. These conditions pose challenges beyond those in controlled laboratory environments. First, specialised systems must coexist and sometimes coordinate with other systems that rely on different inputs, representations, or architectures (Doran et al., 1997; Noura et al., 2019). Second, they may need to coordinate with humans or other agents whose behaviour is neither standardised nor entirely predictable, with shifting goals and action patterns (Dafoe et al., 2020). Third, individual agents and collectives must interact with human users while remaining explainable, controllable, and reliable for oversight (Deroy et al., 2024). Fourth, operating within shared, dynamic spaces places additional demands on artificial systems, which must be selectively responsive to ongoing changes in their environment and to the activities of others in order to act appropriately.

Awareness, as defined above, provides both a practical framework and a design goal for guiding system competence in addressing the challenges of situated interactions, without prematurely invoking the spectre of AI consciousness. Beyond implementation and design, this raises questions about how the awareness of such systems ought to be evaluated.

Here, awareness presents an important epistemic tool: it enables the comparison and evaluation of artificial systems without invoking the categorical divide associated with consciousness. Awareness is readily gradable, allowing systems to be described as more or less aware without conceptual friction. This aligns with calls for ecological and comparative approaches to machine behaviour, which apply methodologies from the quantitative behavioural sciences (Rahwan et al., 2019; Voudouris et al., 2025). While these approaches typically focus on observable behaviour, awareness extends them to internal capacities and system-level characterisations. By foregrounding these practical features, awareness offers a complementary research project, aimed at supporting the evaluation and comparison of systems deployed in dynamic, shared environments.

This paper proceeds as follows. In section 2, we motivate a shift towards the concept of artificial awareness, specifically for the purposes of evaluating and comparing heterogeneous artificial systems in contexts of persistent uncertainty about AI consciousness and the pressing need for evaluation strategies of these system's capacities. In Section 3, we introduce four desiderata that any framework for evaluating awareness in artificial systems should satisfy: it must be domain sensitive, scale-neutral, multidimensional, and ability-oriented. Section 4 unpacks these desiderata and sketches a possible strategy for the practical evaluation of artificial awareness, highlighting provisional dimensions of awareness relevant for embodied artificial systems, and introducing the notion of profiles of awareness for comparative purposes. Finally, in Section 5, we reflect on the remaining conceptual challenges.

## 2. Deserting AI consciousness for artificial awareness

In both public debate and the philosophical literature, the possibility of AI consciousness is hotly debated. Some approach AI consciousness as an inevitable future outcome (Blum & Blum, 2025), others see it as a future possibility given current theories of consciousness (Butlin et al., 2023; Chalmers, 2023), yet others hold it as a possibility but argue it would look very different from consciousness as we know it in humans and other biological organisms (Farisco et al., 2024). These arguments share a reliance on computational functionalism, which holds that phenomenal experience can be realised given a particular implementation of computations. Block (2025) challenged this assumption by introducing what he calls "the meat condition," suggesting that there may be a necessary subcomputational biological condition. Others argue that certain architectural, evolutionary, or developmental conditions may be necessary (Aru et al., 2023) or that consciousness is, and will likely remain, a trait unique to living systems (Seth, 2025).

The broad range of answers available is partially due to confusion about the concept of consciousness itself (Zeman, 2005), the abundance of theories (Kuhn, 2024; Seth & Bayne, 2022), disagreement between core theories (Mudrik et al., 2025a), and the challenges inherent in extrapolating beyond the human case (Negro & Mudrik, 2025). Though we may conquer all these challenges and make progress in consciousness studies, it will take time. Time that, when it comes to AI, we may not have (Schwitzgebel, 2025; Shevlin, 2024).

Though experts may be ambivalent, we need to be concerned that current discussions about the possibility of machine consciousness suffice to influence the public's opinion (Deroy, 2023). A recent study by Colombatto and Fleming (2024) suggests that the idea that large language models (LLMs) can have conscious experiences currently still finds little traction with the public. About a third of participants denied that the LLM could have any experience, and most did not even endorse the description of the system as "somewhat an experiencer." Still, the results also highlight reasons for caution: attributions of experience increase with use of the system, and participants tended to overestimate how willing others would be to attribute consciousness to AI. These attributions of consciousness carry significant weight, influencing people's interactions with these systems, such as their willingness to trust or rely upon them (Colombatto et al., 2025). Moreover, attributions of consciousness may influence how people assign moral status to artificial systems (Gilbert & Martin, 2022), shaping judgments about our ethical responsibilities toward them. Even tentative or partial attributions could therefore carry significant practical and normative consequences.

Most importantly, the growing interest in AI consciousness stems from a legitimate need to understand their ever-expanding capabilities. As systems become more autonomous and are deployed in high-stakes contexts, having principled methods for evaluating their capacities becomes essential for responsible development (Floridi, 2018). The challenge is not whether these capabilities warrant serious evaluation, but rather what conceptual framework best serves that goal. In other words, we are not denying that the motivations behind the discussions about AI consciousness are important. Letting them be framed under the label "consciousness" is misguided. Consciousness brings with it substantial theoretical baggage that may obscure rather than clarify the practical question of what systems can reliably do in context. Moreover, treating consciousness as a binary threshold (present or absent) sits uneasily with the need to compare systems that differ significantly across various axes. Given what is above, the question is whether awareness could do better.

In discussions about biological systems, consciousness and awareness often travel together and are sometimes used interchangeably. Awareness is, in neuroscience, often considered a *component* of consciousness: levels of consciousness (wakefulness) are distinguished from the contents of consciousness (awareness) (Laureys, 2005). Additionally, Block (1995) distinguished between phenomenal consciousness, which refers to the subjective feeling of being in a particular state, and access consciousness, which pertains to the availability of a state's content for reasoning or guiding speech and action. Both distinctions have been influential and subject to sustained debate (Baars & Laureys, 2005; Bayne et al., 2016; Mudrik et al., 2025b).

Outside biological systems, the concept of awareness need not be tied to phenomenal consciousness. It can be understood and applied for practical purposes. A clear example is situation awareness, as formulated by Endsley (1988), which frames awareness in terms of perceiving relevant features of the environment, interpreting their significance, and anticipating potential changes.[1] In aviation, this notion emphasizes how team members coordinate to maintain a shared understanding of the situation (Endsley, 2021). Importantly, this use of awareness is entirely functional: it describes distributed information processing and coordination without making any claims about group consciousness or subjective experience. The concept of situation awareness has also found traction in robotics (Yanco & Drury, 2004).

For artificial systems, we can distinguish between "capacity" and "transient" views of awareness. For capacity views, awareness refers to the artificial realisation of capacities for selectively, processing, storing, and retrieving information in order to guide goal-directed action—encompassing what a system can register as relevant to its operational context and act upon in pursuit of its goals (Evers et al., 2025; Lee et al., 2026). Understood as such, awareness entails a multidimensional space of action-perception abilities, with abilities understood as multi-track dispositions to succeed (Meertens, 2025).

In this respect, the capacity view of awareness has some overlap with Block's (1995) access consciousness, insofar as both emphasise the availability of information for guiding action. This situation of A-consciousness without P-consciousness is often taken to characterise certain non-human animals (Carruthers, 2019), or any creature lacking higher-order mental states (Rosenthal, 2005). However, the similarity is limited. Access consciousness, even when understood functionally, retains a subject-relative character: it describes information that is available to *a subject* for use in reasoning and action. The capacity-view of awareness, by contrast, characterises information processing without reference to a unified subject or integrated perspective, but as a dispositional profile that can be realised across certain contexts in the service of goal-directed action.

---

[1] See also Stanton (2006) for his model of distributed situation awareness as a dynamic collaborative process between agents.

This distinction warrants further unpacking, but it need not be settled here, as there are independent pragmatic grounds for preferring the term awareness over A-consciousness. Keeping "access consciousness" for artificial systems risks backfiring precisely because it carries explicit conceptual entanglements with consciousness debates and equally fraught entanglements with disputes about animal cognition. "Awareness," in this respect, sidesteps both issues without being a term of art or being overtly incomprehensible to the public and legislators.

The transient view goes a step further. Here, attributing awareness to a system operating within a specific domain entails a claim about *a state* at time t: the system processes and utilizes information in a situated and integrated way within this narrow context. Crucially, this approach remains agnostic about whether such states imply continuous awareness or merely transient episodes. For artificial systems, awareness need not correspond to a unified, ongoing perspective or a standing capacity; instead, it suffices to identify transient realisations of awareness that arise under particular conditions (see also Deroy et al., 2024). The emphasis would then be on measuring the states a system can enter, rather than capturing continuous dispositions.

Throughout this paper, we adopt the capacity view of awareness, leaving a more detailed philosophical discussion of the transient view for another occasion. However, regardless of the view adopted, the shift from consciousness to awareness provides a promising foundation for a comparative, operational tool for evaluating artificial systems. This is particularly salient given the epistemic situation presented by artificial systems. Unlike biological organisms whose consciousness we attempt to discover, artificial systems are designed to meet specified functional demands (Simon, 1996). The question thus shifts from asking whether a system's existing capacities resemble those observed in humans to the more practical inquiry: "What needs to be there for the system to do x?"

## 3. Desiderata for evaluating awareness

Artificial systems present a wide range of variations. A swarm of firefighting drones is significantly different from an image-based neural network used for medical diagnosis. This poses a central challenge for evaluation. Both systems might reasonably be described as aware, yet it remains unclear how to go about comparing the differences and similarities between them. What form of evaluation could be applied to both in a *fair* and *structured* manner?

We identify four desiderata for evaluating artificial awareness: evaluation must be domain-sensitive, multidimensional, deployable across scales, and capable of predicting task performance while allowing for generalisation at the level of abilities.

## 3.1 The evaluation must be domain-sensitive

Since its founding, a driving force behind AI research has been the quest for Artificial General Intelligence, or AGI (Goertzel, 2014). While definitions vary, AGI typically refers to a hypothetical future artificial system capable of performing at or beyond human levels across a wide range of tasks (Shevlin, 2020). This relies on the notion of general intelligence as the ability of an agent to utilise information in achieving diverse goals under various conditions (Legg & Hutter, 2007). Today's AI falls short of the intelligence found in biological organisms, let alone humans (Shevlin et al., 2019); most current systems are specialised (Roli et al., 2022).

Opinions diverge on what achieving AGI would require. Most emphasize the need to pursue varied goals across contexts, find novel solutions to problems, and generalise and transfer skills across domains (Goertzel, 2014). However, some argue that insight (Roitblat, 2020), or biological agency (Roli et al., 2022) would be needed. Others bring this debate back to consciousness (Juliani et al., 2022). Independently of the challenges of realising AGI, there are substantive reasons to question both the necessity and desirability of developing artificial systems endowed with general intelligence or human-level abilities across domains (Deroy et al., 2024).

One reason concerns sustainability. Domain-general architectures are typically resource-intensive in manufacturing, training, and deployment, and can consume substantial energy even when applied to narrowly defined tasks (Wu et al., 2022). Using a specialised system is often more efficient, especially at scale (Luccioni et al., 2024). A second reason concerns oversight and control. Systems designed for specific domains are easier to explain, monitor, and regulate because their operational scope and performance criteria are constrained. By contrast, domain-general systems risk greater opacity, complexity, and behavioural unpredictability. A third reason is that the goal of achieving "human-level" performance unfairly presupposes a single dimension of analysis (McDermott, 2007).

These considerations bear directly on evaluation. If narrow AI agents constitute our primary targets of assessment, at least for the near future, this should shape how we evaluate their competencies. A swarm of delivery robots and a large language model operate in vastly different domains, with different affordances and constraints on performance. A too rigidly determined set of benchmarks or tests is unlikely to provide a meaningful assessment for both.

This mirrors existing challenges in comparative cognition: assessing awareness in corvids versus cephalopods requires different considerations. A highly social bird faces radically different informational demands and adaptive pressures than a primarily solitary marine invertebrate (Emery & Clayton, 2004; Godfrey-Smith, 2018). Their motor capacities, social environments, and survival challenges differ so greatly that the relevant information, adaptive responses, and measures of success diverge. This need not preclude meaningful comparison: both could exhibit similar awareness profiles, be *similarly responsive* to types of information relevant within their domains, albeit through different mechanisms. Any evaluation framework must therefore account for such specialisations and be sensitive to the "ecological niche" of the system.

## 3.2 The evaluation must be multidimensional

The diversity of systems under consideration brings us to a familiar methodological challenge: how to evaluate similarities and differences without collapsing meaningful variation. Any framework must strike a balance between abstraction and specificity, capturing relevant distinctions while remaining sensitive to commonalities. Research on disorders of consciousness has increasingly abandoned hierarchical models, arguing that global states of consciousness cannot be ordered along a single dimension (Bayne et al., 2016; Bayne & Carter, 2018), and instead appealing to multidimensional approaches to capture finer-grained differences. Similar concerns have motivated multidimensional accounts of animal consciousness (Birch, 2020; Dung & Newen, 2023; Veit, 2023) and proposals for AI consciousness (Evers et al., 2025).

A multidimensional approach allows evaluation to reflect both the independence and interdependence of different aspects of awareness. Rather than asking whether a system exhibits more or less awareness overall, it enables comparisons across distinct dimensions. Applied to artificial systems, this lets us assess capabilities independently, recognizing that a system may exhibit sophisticated awareness in one dimension while showing minimal capacity in others. This is especially important given the specialised nature of many artificial systems, which, though restricted to specific domains, can still vary along different dimensions of awareness.

## 3.3 The evaluation must be deployable at different scales

Contemporary artificial systems frequently rely on distributed, modular, or multi-agent architectures (Balaji & Srinivasan, 2010; Tran et al., 2025). Combinations of LLMs with Mixture of Experts frameworks leverage the cooperation of various deep learning models to improve efficiency and optimization (Du et al., 2024; Shazeer et al., 2017). Swarm robotics draws on the behaviours of social animals, such as eusocial insects and bird flocks, to generate robust, scalable, and flexible behaviours at the collective level by implementing simple rules at the level of individuals. The collective behaviour emerges through the local interactions between robots (Brambilla et al., 2013; Garnier et al., 2007). These cases raise a question that has so far received surprisingly little attention in the philosophical literature: what is the appropriate scale for evaluation?

Collective frameworks have been gaining traction in adjacent debates. Some have argued that artificial systems can be understood as collective or group agents and legislated accordingly, similar to group agents such as corporations or corporate subcontractors (List, 2021; Jennings & Montemayor, 2025). Others apply the lens of collective agency to human-machine cooperation and multi-agent systems (Misselhorn, 2015). A parallel in biology is the treatment of eusocial insect colonies as "superorganisms" (Hölldobler & Wilson, 2009). Huebner (2014) similarly argues that some collectives should be viewed as intentional systems capable of exhibiting a collective mentality. Nevertheless, there remains little guidance on when it is appropriate to adopt individual versus collective frameworks for evaluating the abilities of artificial systems.

The ambiguity here mirrors the special composition question in metaphysics (van Inwagen, 1987): under what conditions do multiple entities compose a single unified whole? This question is especially pressing for multi-agent or distributed systems where collective behaviours emerge from the interactions of single units. Luckily, we do not need to resolve this presently. Instead, it highlights a limitation in frameworks that presuppose the correct scale of analysis a priori. Evaluating individual drones tells us little about a swarm's capabilities, as assessing components in isolation may miss emergent system-level competencies. Conversely, treating every collection of AI components as a unified agent may obscure important differences. Any attempt to evaluate both such systems fairly, then, does not presuppose the individual as the correct scale but instead remains neutral at the outset.

### 3.4 The evaluation must predict task performance whilst generalising to abilities

A key problem faced by animals and artificial systems alike is that we primarily rely on observed behaviour to evaluate them. At the same time, the role of evaluation is not to provide direct access to internal states, but to give structured evidence from which states could be reasonably inferred. Exactly what evidence justifies such inferences remains to be seen. A related challenge arises from the diversity of organisms and systems under study: comparing very different animals or artificial systems raises methodological issues. These shared challenges have led some to argue that AI research should draw on comparative psychology and cognition for methods and measures (Buckner, 2023; Shevlin & Halina, 2019). For instance, psychometric tests (Hernández-Orallo, 2017b) or tests of animal consciousness (Dung, 2023) have been highlighted, including testbeds such as the "Animal-AI Olympics" (Crosby et al., 2019; Crosby, 2020). Relatedly, similar biases, such as anthropomorphism (Barrett, 2016), are likely to influence assessments.

Both fields must therefore develop methods to interpret success and failure in task performance fairly. Firestone (2020) relates this to the distinction between competence and performance: competence refers to the underlying knowledge a system has, whereas performance entails its use of this competence in a single instance. Performance may fail to accurately reflect competence due to constraints, and success may occur despite the absence of underlying competence. To facilitate a species-fair comparison, Firestone argues that the particular performance constraints of machines need to be taken into account in their evaluation.

A related distinction concerns task-oriented versus ability-oriented evaluation. Hernández-Orallo (2017a) argues that specialised AI systems are most appropriately evaluated at the level of tasks, whereas more general systems call for ability-oriented evaluation. The reasoning is that, given the narrow scope of specialised systems, their evaluation can be specified relative to their concrete context. Ability-oriented evaluation, by contrast, becomes relevant once systems operate across multiple contexts or applications and must rely on more general capacities, such as reasoning or inductive learning.

While intuitively appealing, this division does not map neatly onto artificial systems in practice. Many specialised systems operate in dynamic environments, interact with other agents, and must cope with uncertainty and change. In such settings, successful task performance already presupposes more generalised abilities. There will be cases in which a system's task environment is sufficiently constrained that no appeal to such abilities is required. However, this cannot be assumed to be the case for all domain-specific systems at the outset. Treating task-level evaluation as sufficient by default risks focusing evaluation on isolated successes rather than on the competences that enable a system to cope with variation, uncertainty, and change. Where evaluation is intended to inform ethical judgment, oversight, or human–AI collaboration, such ability-level information is essential, as it bears directly on how the system is likely to behave outside narrowly controlled conditions, or when things go wrong.

The notion of abilities we use here draws on the distinction between generic and specific abilities (Mandelkern et al., 2017). Generic abilities are not time-indexed; for example, the general ability to draw a house, rather than to draw one right now. Applied to machine learning, Harding and Sharadin (2024) define a model as having a capability to φ if, under a set of appropriate conditions, it would reliably succeed at φ when it "tries." They highlight three features: reliability, competence versus performance, and non-coincidence. These address earlier challenges: reliability distinguishes genuine abilities from isolated successes, the competence-performance distinction accounts for contextual constraints, and non-coincidence prevents mistaking memorisation or gaming for underlying competence.

Thus, evaluating a system's awareness fairly requires focusing on its abilities rather than isolated task performance, and on designing tasks that are appropriate for the system and the environment in which it operates. This helps avoid both over-attributing abilities due to instances of gaming (Dung, 2023) and under-attributing abilities by failing to consider system-specific constraints, whilst making both differences and similarities between systems visible.

## 4. Structuring the evaluation of artificial awareness

If awareness is to play a meaningful role as a property of artificial systems, supporting their operation in dynamic, shared environments, then it must be susceptible to principled evaluation. Artificial agents are increasingly interacting with humans and with other artificial systems that differ in architecture, scale, and purpose. In such settings awareness can underwrite coordination, adaptation, and goal-directed action beyond narrowly constrained situations. Treating awareness as a target of evaluation allows us to assess not only what a system can do in isolation, but how its action–perception abilities combine, interact, and remain robust across changing conditions, as well as the scope and limits of its capacities relative to designed tasks. Evaluation is therefore central to system design, oversight, and responsible integration. The question, then, is how awareness can be evaluated across artificial systems, and how the desiderata outlined above can be translated into a concrete evaluative approach.

The discussion proceeds by outlining key features for awareness evaluation grounded in the earlier desiderata (Section 4.1), examining five provisional dimensions—spatial, temporal, metacognitive, agentive, and self-awareness (Section 2.2)—and exploring how dimensional profiles support systematic comparison within and across artificial systems (Section 4.3).

### 4.1 From desiderata to evaluation

The desiderata established in the previous section jointly motivate the claim that the assessment of awareness must be appropriate to the situations in which a system is deployed. Evaluation should therefore be sensitive to domain-specific demands, as well as to the system's own constraints and affordances. Against this background, rather than offering a dispositive account, we propose conceptual scaffolding for evaluating and comparing awareness in artificial systems, leaving a fully developed and implementable framework for future work.

To realise this, we draw on prior work by Lee et al. (2026), who outline a multidimensional approach to measuring the impact of awareness on system performance. Their focus lies on the practical measurability of awareness; in essence, they are concerned with which action-perception abilities a system—concretely, a robot swarm—should possess to improve performance. They introduce a pipeline that begins with establishing the dimension of awareness of interest, then proceeds through the associated capacities, the mechanisms that enable these capacities, evaluation tasks, and finally performance metrics. Crucially, their approach assumes awareness is present and then asks how this impacts system performance in a given task. Here, we expand upon these elements to outline an approach to evaluating whether awareness is present in the first place.

In what follows, we distinguish three related elements: (i) dimensions of awareness, (ii) action-perception abilities, and (iii) the tasks in which such abilities are expressed and assessed.

**Dimensions of awareness.** A key distinction can be drawn between dimensions as understood in mathematics or physics (Sterrett, 2021) and the conceptual or evaluative understanding of dimensions often used in the philosophy of mind and psychology. In the latter sense, dimensions function not as formal parameters within a quantitative space, but as conceptual categories used to structure comparison and explanation.[2] Understood in this light, conceptual dimensions exhibit several key characteristics. First, they are gradable: variation along each dimension can, in principle, be assessed in terms of degrees. Second, they display conceptual distinctness: dimensions must be sufficiently differentiated to warrant separate evaluation, even if they turn out to be empirically related. Third, they provide explanatory utility, enabling informative comparisons across systems and supporting higher-level mappings of cognitive or experiential capacities.

A further consideration is the distinction between separable and integral dimensions, from psychophysics (Garner & Felfoldy, 1970). Separable dimensions can be analysed independently, as with colour and shape, while integral dimensions are co-instantiated, such as hue and saturation. Although separable dimensions are simpler to work with, integral dimensions would prove theoretically informative: in our context, such a finding would indicate systematic interdependence between aspects of awareness.

---

[2] For instance, Birch et al. (2020) propose a dimensional framework for animal consciousness, including perceptual richness, evaluative richness, unity, temporality, and selfhood. These dimensions are intended to capture fine-grained variation in conscious capacities across species and to facilitate systematic, comparative research. This framework illustrates the kind of conceptual structuring that motivates our use of dimensions in evaluating artificial awareness.

**Action-perception abilities.** Abilities represent the primary locus of evaluation in our framework. Following Harding and Sharadin's (2024) approach to abilities, we understand abilities as general capacities that manifest across varied conditions and contexts, rather than isolated performances or one-off successes.

An ability, in this sense, reflects a latent dispositional construct of what an agent or collective can do, where "can do" is understood in terms of *reliable* and *non-coincidental* success when attempting *relevant* tasks across a range of conditions. Consider spatial navigation as an example: an agent with robust spatial awareness abilities does not merely succeed at navigating from point A to point B on a single occasion, but demonstrates consistent success across varied environments, lighting conditions, obstacle configurations, and task demands. Abilities are gradable in two respects. First, systems may possess more or fewer distinct abilities along any given dimension. Second, individual abilities vary along a range of metrics.

This understanding of abilities aligns with broader philosophical work on skills and competences, where the possession of an ability is distinguished from its exercise, and where abilities are understood as involving dispositions or know-how that generalises across instances (Fridland, 2014; Ryle, 1949 / 2007; Vetter, 2019). Critically, abilities in artificial systems are constituted by action-perception couplings—they reflect not passive information processing but the capacity to use perceptual and stored information to guide goal-directed action or decision. This active character is what makes abilities the appropriate target for evaluating awareness in artificial systems, which are ultimately assessed by what they can accomplish in interaction with their environments.

**Evaluation tasks.** While abilities provide stable points of comparison across systems, tasks must be adapted to accommodate the specific constraints and affordances of each system under evaluation (cf. Dung & Newen, 2023, on species-sensitive validation). This approach draws on established practices in comparative cognition, where researchers evaluate similar abilities across species using tasks adapted to each species' sensorimotor capabilities and ecological niche. For instance, the string-pulling paradigm (Jacobs & Osvath, 2015), the detour paradigm (Kabadayi et al., 2017), or delayed gratification (Miller et al., 2019) have been successfully adapted across a range of taxa. The key is that while the surface features of tasks vary with implementation constraints, they are intended to target the same abilities.[3]

Critically, evaluation requires a battery of tasks rather than single measures. This is necessary for two reasons. First, establishing that a system possesses an ability requires demonstrating success across varied conditions and task instantiations. Second, characterising the degree to which a system possesses an ability requires assessing it along multiple **performance metric**s, such as reliability (consistency across trials), robustness (performance under perturbation or degradation), and flexibility (generalisation to novel conditions). A comprehensive task battery thus serves both to establish whether an ability is present and to map its scope and limitations.

---

[3] There is ongoing debate in comparative cognition about whether these paradigms do indeed succeed in reliably measuring the same underlying abilities across species. For discussion of these methodological and interpretive challenges, see Shettleworth (2010) Penn and Povinelli (2007).

This approach prevents both false positives (mistaking narrow task-specific competence for genuine ability) and false negatives (failing to detect genuine abilities due to reliance on single or poorly adapted tasks).

Taken together, this yields a three-part evaluative structure: dimensions that carve the space of awareness into distinct conceptual domains, abilities that pick out stable action–perception competencies within those domains, and tasks that probe these abilities in species- and system-sensitive ways. This structure provides a way of realising the desiderata outlined above, supporting domain-sensitive and multidimensional comparison while remaining scale-neutral and oriented toward abilities rather than isolated performances.

## 4.2 Dimensions of awareness

For clarity's sake, we now illustrate one potential avenue towards realising such a situated approach to evaluating artificial awareness, which is in line with this three-part structure.

We identify five provisional dimensions of awareness: spatial, temporal, metacognitive, agentive, and self-awareness. These correspond to distinct informational domains that systems may exploit in guiding goal-directed action. While emphasised here for their relevance to embodied artificial systems, other AI applications might prioritise different dimensions. These dimensions are not intended to exhaust the space of possible dimensions or serve as a definitive taxonomy; instead, they capture a broad range of informational sensitivities that can be meaningfully varied and assessed across diverse artificial systems.

**Spatial awareness**: refers to a system's abilities to detect, differentiate, and exploit spatial relations—such as distance, direction, proximity, and orientation—among relevant elements of its environment in ways that guide its actions. For artificial systems, this involves abilities such as navigation, localisation, mapping, or tracking. At a basic level, artificial systems can use GPS to track their location. Beyond this, robots may employ feature-based localization, identifying and mapping specific environmental landmarks (e.g., Jones & Hauert, 2023). Or approaches involving Simultaneous Localization and Mapping (SLAM), which enables a system to build a map of its surroundings while determining its own position within it (Taheri & Xia, 2021).

**Temporal awareness**: refers to a system's abilities to detect, differentiate, and exploit temporal relations—such as continuity, duration, simultaneity, persistence, change, and succession—among relevant elements of its environment in ways that guide its actions. For artificial systems, this involves abilities such as sequencing, prediction, anticipation, or temporal coordination. In biological organisms, the ability to perceive and process temporal information is central to cognition (Buhusi & Meck, 2005), playing a key role in processes such as conditioning and reinforcement learning (Gallistel & Gibbon, 2000). Temporal perception is also important in embodied and bio-inspired robotics (Maniadakis et al., 2009), as well as in multi-agent systems, including human-robot interaction (Maniadakis et al., 2020) and teams composed of heterogeneous agents (Maniadakis et al., 2016).

**Self-awareness**: minimally understood, refers to a system's ability to monitor and respond to its own *physical* states. Although self-awareness is often discussed in rich phenomenological terms (Grünbaum & Zahavi, 2013; Nida-Rümelin, 2016), here we focus on its minimal instantiation—bodily awareness. This includes abilities such as interoception, proprioception, and exteroception. In artificial systems, a key aspect of bodily awareness is error or fault recognition, where the system detects discrepancies between its expected and actual states, such as malfunctions or damage. A practical application of bodily awareness can be seen in soft robotics: Soter et al. (2018) integrate exteroceptive and proprioceptive sensors, enabling a robot to "imagine" its own position even in the absence of visual input.

**Metacognitive awareness**: refers to a system's abilities to monitor, assess, and regulate its own processing to guide its actions. For artificial systems, this could involve abilities such as evaluating their confidence, adjusting strategies based on feedback, or determining when to seek additional information or opt out. Beyond such self-directed abilities, there are also related other-directed capacities to consider, such as those involved in mentalising or mind-reading (Proust, 2014). In humans, the ability to monitor and respond to the mental states of others supports coordination and collaboration (Frith, 2012). Similarly, metacognitive capacities may contribute to improved safety, robustness, and coordination in artificial systems, particularly in human–machine interaction contexts (Chiba & Krichmar, 2020; Johnson, 2022; Mörwald, 2011, Winfield, 2018).

**Agentive awareness:** refers to a system's abilities to identify, evaluate, and modulate its own actions in relation to the affordances and constraints presented by its environment, itself, and other agents. For artificial systems, this involves abilities such as recognising action possibilities in dynamic contexts, coordinating with others, or adapting motor strategies based on changes in the environment. An affordance-based perspective, rooted in ecological psychology (Gibson, 1979), has already found practical application in robotics, where systems are being designed to dynamically perceive and respond to action possibilities within their operational contexts (Jamone et al., 2018; Ardón et al., 2020).

Although only a preliminary sketch, these five dimensions satisfy the criteria outlined earlier: they admit graded variation, are conceptually distinct, and have the potential to capture informative differences in awareness across artificial systems, insofar as each tracks a different informational domain that may be exploited in action–perception coupling. While the dimensions are conceptually separable, they are expected to overlap in practice.[4] Tracking an object moving through space, for instance, typically involves sensitivity to both its spatial location and the temporal unfolding of its movement. Spatial and temporal awareness, therefore, often co-occur; yet, they can be separated. A neural network, for example, may be trained to detect duration without any spatial input at all.

---

[4] The five dimensions identified here should be understood as provisional. We leave for future investigation questions concerning their interdependencies, their relationship to consciousness, whether some dimensions are more fundamental than others, and whether this set admits of further refinement through decomposition or hierarchical organization.

At this stage, we deliberately leave the specification of concrete tasks and metrics open. The aim is not to provide a universal, one-size-fits-all assessment battery, but to offer conceptual scaffolding—a recipe that researchers and developers can adapt to particular systems, domains, and evaluative aims. The relevant abilities, how they should be measured, and the appropriate experimental paradigms will vary substantially between, for example, a warehouse logistics robot, a surgical assistant, and a domestic companion system. The salience of particular dimensions likewise shifts across system types. Conversely, dimensions that are peripheral for some systems may become central for others, such as social or cultural dimensions of awareness in the case of LLMs (Li et al., 2024).

## 4.3 Profiles of awareness

A core motivation for this approach to awareness is to enable structured evaluation and comparison of artificial systems. Not just for its own sake: more precise characterisation and description of the abilities of artificial systems are intended to improve communication between AI developers and end-users, avoid miscommunication with the public, support more informed oversight, and targeted system design. An open question, then, is how the picture of awareness painted so far might enable such comparisons in practice. To answer this, we draw inspiration from the notion of profiles of consciousness, as employed by Dung and Newen, and instead leverage profiles of awareness (cf. Birch et al., 2020; Dung & Newen, 2023; Evers et al., 2025).

Four types of comparison are at stake, each with distinct challenges and benefits. First, a comparison of a single artificial system with varying configurations. Here, the emphasis lies on evaluation for improving system performance. Consider an AI application tested on a standard task without awareness capabilities, then retested with various awareness abilities variably enabled or disabled. Each combination generates a distinct awareness profile that can inform the optimal configuration, given the system's role. In a recent study, Lee et al. (2026) tested a robot swarm in an intralogistics scenario with certain abilities present or absent, specifically examining two levels of spatial awareness (knowledge of box quadrant versus exact location) and two levels of bodily awareness (fault detection versus fault diagnosis). Interestingly, the highest levels of information along both dimensions did not combine to yield the best performance in terms of efficiency. Having access to more information need not imply better performance; in some circumstances, being just aware enough may be preferable.

Second, comparisons between congruent artificial systems. These comparisons are relatively straightforward insofar as systems operating on the same or similar architectures are likely to face comparable constraints and opportunities. Even when applied in different domains, such systems' awareness abilities can likely be evaluated using similar task designs with only minimal system-specific adjustments.

Third, comparisons between heterogeneous artificial systems. These comparisons are inherently more speculative, as they will either yield more general, less detailed insights or require adopting experimental paradigms typically suited for cross-species comparisons. To facilitate fair evaluation, parameters must be equalized: the same abilities need to be assessed using a battery of tests, and the same metrics must be applied across systems. The same paradigms can be employed across different systems, though their concrete implementation will require adaptation to accommodate differences, preserving the underlying functional principle while adjusting surface-level task features. Figure 1 presents hypothetical profiles for three diverse systems (a delivery robot swarm, an agricultural gripper, and a medical image classification network), illustrating how such comparisons make visible both commonalities and differences across architectures.

Fourth, comparisons between artificial systems and biological organisms, such as human or non-human animals. These comparisons present the most significant methodological challenges due to fundamental differences in substrate, evolutionary history, and architecture. Significant questions remain about how to realise such a comparison in practice, especially as this brings the debates about consciousness (and its potential intersections with awareness as described here) back into the picture.

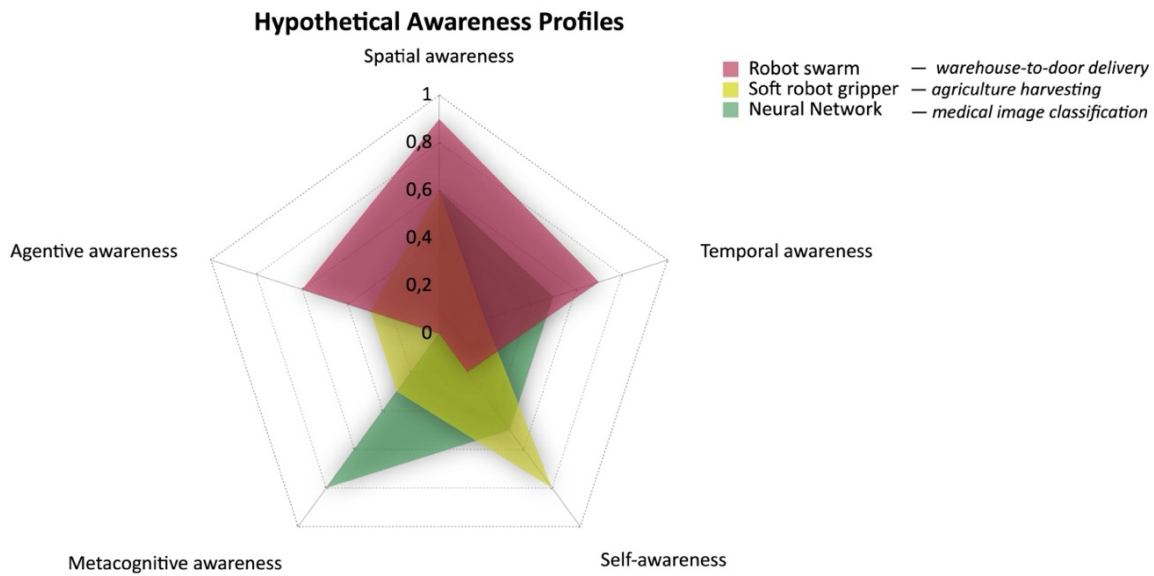

*Figure 1: Illustrative awareness profiles across three diverse artificial systems. Each dimension is assessed independently, revealing distinct patterns of capabilities suited to different operational domains.*

## 5. Back to the initial question: When is awareness useful?

We can now revist a point raised earlier. Why do we need to reference awareness to account for these abilities artificial systems have to adapt to their environment and coordinate with others?

Indeed, various other strategies have been employed to meet these demands. These approaches can schematically be positioned along a spectrum. At one end are rule-based or heuristic approaches that aim to tackle the interaction problem without appealing to cognitive, representational, or mentalistic concepts (Brooks, 1991; Dreyfuss, 1992; Pollack, 2014). These approaches aim to produce adaptive behaviour through simple reactive mechanisms or carefully engineered stimulus-response patterns. At the other end are approaches that invoke neuromorphic architectures, aiming to support phenomenal consciousness (Schneider, 2019; Verschure, 2016), treating subjective experience as central to genuine flexibility and adaptation.

Between these extremes lies a range of intermediate views, one of which is captured by the concept of awareness. Designing aware systems marks an intermediate path. It frames a largely empirical research programme concerned with whether artificial systems can take up and use information about their own processing, about other agents, or action possibilities in their surroundings in sufficiently flexible and systemic ways. The value of this framing lies in offering a plausible path to develop and understand flexible interactive systems while not assuming present or future consciousness.

Crucially, not every system exhibiting goal-directed information processing counts as aware on this account. As developed through the ability-centred approach (see section 3.4), awareness requires reliable, non-coincidental abilities that generalise across appropriate variations in context, rather than brittle, coincidental successes. A thermostat that is sensitive to temperature but cannot adapt this response, generalise across conditions, or integrate this information with other sources—lacks the dispositional profile that characterises awareness. The account we present here identifies which systems possess these richer capacities and to what degree, while leaving metaphysical questions about awareness's ultimate nature for future inquiry.

Another open point concerns the apparent proximity of this evaluation approach to behaviourist accounts. Awareness, as characterised here, is indeed evaluated primarily in terms of observable behaviours, rather than by appeal to specific internal mechanisms or processing architectures. This is primarily a methodological choice. The aim is not to deny the importance of mechanisms, but to avoid assuming which mechanisms are necessary at the outset, especially given that their realisation is likely to vary across systems.

That is not to say that mechanistic details are not crucial for implementation and evaluation, for instance, in experimental paradigms where specific abilities can be selectively turned on or off to assess their contribution to performance (Lee et al., 2026). The framework, therefore, separates criteria for attributing awareness from claims about its underlying realisation, allowing comparison across systems without prematurely constraining how awareness must be implemented, while recognising that architectural considerations will ultimately matter (Wong, 2025).

## 6. Conclusion

This paper was motivated by the question of whether a repositioned notion of awareness can better address challenges currently grouped under the heading of AI consciousness, and how awareness can be evaluated across artificial systems. In response, we outline a complementary program, a pragmatic lens, aimed at developing principled methods for evaluating and mapping the capacities of artificial systems. We argue that *awareness*—understood as a system's ability to take up information from its surroundings, process it, and act upon it in a goal-directed way—supports such evaluations while carrying less metaphysical and ethical baggage than consciousness, thereby enabling more productive scientific and public discourse. Beyond its evaluative role, treating awareness as a design goal may also help address practical challenges faced by AI operating in dynamic, shared environments, facilitating coexistence and cooperation with other artificial and human agents, while supporting meaningful forms of oversight and control.

As a first step toward these goals, this paper outlined the preliminaries of an approach to evaluating awareness and identified a set of core desiderata: evaluations should be domain-sensitive, multidimensional, deployable across scales, and capable of predicting task performance while allowing for generalisation at the level of abilities. These desiderata were operationalised within a structured comparative framework comprising three features: dimensions, abilities, and tasks. The dimensional scheme proposed here offers one possible instantiation of this approach, while leaving room for further refinement and empirical development. In essence, we think that a focus on awareness can shift the conversation—from whether AI can do everything humans can do, to asking what we actually *need* artificial systems to do, and which capacities are required for them to be *just aware enough* to succeed at those tasks.